# Chapter 5: Foundations of Data Imbalance and Solutions for a Data Democracy

Ajay Kulkarni, Feras A. Batarseh, and Deri Chong

*"In the end, it's all a question of balance"* – Rohinton Mistry

**Abstract**

Dealing with imbalanced data is a prevalent problem while performing classification on the datasets. Many times, this problem contributes to bias while making decisions or implementing policies. Thus, it is vital to understand the factors which causes imbalance in the data (or class imbalance). Such hidden biases and imbalances can lead to data tyranny, and a major challenge to a data democracy. In this chapter, two essential statistical elements are resolved: the degree of class imbalance and the complexity of the concept, solving such issues helps in building the foundations of a data democracy. Further, statistical measures which are appropriate in these scenarios are discussed and implemented on a real-life dataset (car insurance claims). In the end, popular data-level methods such as Random Oversampling, Random Undersampling, SMOTE, Tomek Link, and others are implemented in Python, and their performance is compared.

## 1. Motivation & Introduction

In the real-world, data are collected from various sources like social networks, websites, logs, and databases. Whilst dealing with data from different sources, it is very crucial to check the quality of the data [1]. Data with questionable quality can introduce different types of biases in various stages of the data science lifecycle. These biases sometime can affect the association between variables, and in many cases could represent the opposite of the actual behavior [2]. One of the causes that introduce bias in the interpretation of results is data imbalance. A problem that especially occurs while performing classification [3]. It is also noted that most of the time data suffer from the problem of imbalance which means that one of the classes has a higher percentage compared to the percentage of another class [4]. In simple words, ***a dataset with unequal class distribution is defined as imbalanced dataset [5]***. This issue is widespread, especially in binary (or a two-class) classification problems. In such scenarios, ***the class which has majority instances is considered as a majority class or a negative class, and the underrepresented class is viewed as a minority class or a positive class***. These kinds of datasets create difficulties in information retrieval and filtering tasks [6] which ultimately results in a poor quality of the associated data. Many researchers studied similar applications of class imbalance, such as fraudulent telephone calls [7], telecommunications management [8], text classification [4][9][10][11], and detection of oil spills in satellite images [12].

The machine learning/data mining algorithms for classification are built on two assumptions: *Maximizing output accuracy, and test data is drawn from the same distribution as the training data*. In the case of imbalanced data, one or both the assumptions get violated [13]. Let's consider the example of fraud detection to understand the issue of imbalanced data (or class imbalance) more clearly. Suppose there is a classifier system which predicts fraudulent transactions based on the provided data. In real life, there will always be less percentage of people who will do fraudulent transactions, and most of the instances in the data will be of non-fraudulent transactions. In that case, the given classifier will always treat many of the fraudulent transactions as non-fraudulent transactions because of the unequal percentage of class distribution in the data. The classifier system will result in a high percentage accuracy but a poor performance. Therefore, it is critical to deal with this issue and build accurate systems before focusing on the selection of classification methods.

The purpose of this chapter is to understand and implement common approaches to solve this issue. The next section of this chapter focusses on the causes of imbalanced data and how accuracy (and traditional quality measures) can be misleading metric in these cases. After that, different solutions to tackle this issue are discussed and implemented.

## 2. Imbalanced Data Basics

The previous section introduced the meaning of positive class, negative class and the need to deal with imbalanced data. In this section, the focus will be on the factors which create difficulties in analyzing the imbalanced dataset. Based on the research of Japkowicz et al. [14], the imbalance problem is dependent on four factors: ***degree of class imbalance, the complexity of the concept represented by the data, an overall size of the training data, and type of the classifier***. Japkowicz et al. [14] conducted experiments by using sampling techniques and then compared results using different classification techniques to evaluate the effects of class imbalance. These experiments indicated the importance of complexity of the concept and reflected that the classification of a domain would not be affected by huge class imbalance if the concept is easy to learn. Therefore, the focus of the solution ought to be to understand two essential factors - ***degree of class imbalance*** and ***complexity of the concept***.

### 1) Degree of class imbalance

The degree of class imbalance can be represented by defining the ratio of positive class to the negative class. There is also another way to understand the degree of imbalance by calculating the Imbalanced Ratio (IR) [5], for which the formula is given below:

$$IR = \frac{Total\ number\ negative\ class\ examples}{Total\ number\ of\ positive\ class\ examples}$$

Let's say if there are 5000 examples that belong to a negative class and 1000 examples that belong to a positive class, then we can denote the degree of class imbalance as 1:5; i.e., there is a positive class example for every five negative class examples. Using the above formula, *Imbalanced Ratio (IR)* can also be calculated as follows:

$$IR = \frac{Total\ number\ negative\ class\ examples}{Total\ number\ of\ positive\ class\ examples} = \frac{5000}{1000} = 5$$

Thus, *IR* for the given dataset would be 5. In this way, the degree of class imbalance can provide information about data imbalance and can help structure the strategy for dealing with it.

### 2) Complexity of the concept

The complexity of the concept can be majorly affected by class overlap and small disjoints. *Class overlap occurs when examples from both the classes are mixed at some degree in the feature space* [5]. Figure 1(a) shows the class overlap in which circles represents a negative class and triangles represent a positive class.

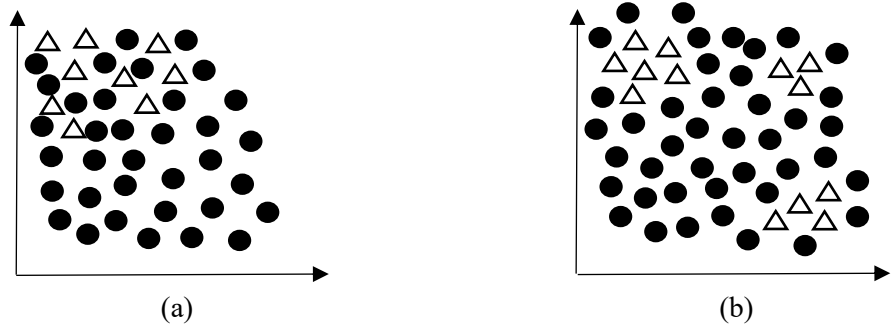

**Figure 1:** Example of (a) Class Overlap and (b) Small Disjuncts

Another factor which affects the complexity of the concept is small disjoints which are shown in Figure 1 (b). Small disjoints occur when the concept represented by the minority class is formed of sub-concepts [15] [5]. It can be easily seen from the figure that class overlap and small disjoints introduce more complexity in the system, which affects class separability. This results in more complex classification rules and ultimately misclassification of the positive class instances.

Now let's focus on different approaches which can help to deal with these imbalance issues. The presented approaches (in the next section) help to improving the quality of the data for better analysis and improved overall results for data science.

## 3. Statistical Assessment Metrics

This section outlines different statistical assessment metrics and various approaches to handle imbalanced data. To understand this section, ***"Porto Seguro's Safe Driver Prediction"*** [16] dataset is used for implementing various techniques available in Python. This section starts with statistical assessment metrics that provide insights into classification models.

### 3.1 Confusion Matrix

Confusion matrix is a very popular measure used while solving classification problems. It can be applied to binary classification as well as for multi-class classification problems. An example of a confusion matrix for binary classification is shown in Table 1.

|  |  | **Predicted** | |
|---|---|---|---|
|  |  | **Negative** | **Positive** |
| **Actual** | **Negative** | TN | FP |
|  | **Positive** | FN | TP |

**Table 1:** Confusion Matrix for Binary Classification

Confusion matrices represent counts from predicted and actual values. The output "TN" stand for True Negative which shows the number of negative examples classified accurately. Similarly, "TP" stands for True Positive which indicates the number of positive examples classified accurately. The term "FP" shows False Positive value, i.e., the number of actual negative examples classified as positive; and "FN" means a False Negative value which is the number of actual positive examples classified as negative. One of the most commonly used metrics while performing classification is accuracy. The accuracy of a model (through a confusion matrix) is calculated using the given formula below.

$$Accuracy = \frac{TN + TP}{TN + FP + FN + TP}$$

Accuracy can be misleading if used with imbalanced datasets, and therefore there are other metrics based on confusion matrix which can be useful for evaluating performance. In Python, confusion matrix can be obtained using *"confusion_matrix()"* function which is a part of *"sklearn"* library [17]. This function can be imported into Python using *"from sklearn.metrics import confusion_matrix"*. To obtain confusion matrix, users need to provide actual values and predicted values to the function.

### 3.2 Precision and Recall

Precision and recall are widely used and popular metrics for classification. Precision shows how accurate the model is for predicting positive values. Thus, it measures the accuracy of a predicted positive outcome [18]. It is also known as the Positive Predictive Value. Recall is useful to measure the strength of a model to predict positive outcomes [18], and it is also known as the Sensitivity of a model. Both the measures provide valuable information, but the objective is to improve recall without affecting the precision [3]. Precision and Recall values can be calculated in Python using "precision_score()" and "recall_score()" functions respectively. Both of these functions can be imported from "sklearn.metrics" [17]. The formulas for calculating precision and recall are given below.

$$Precision = \frac{TP}{TP + FP} \qquad Recall = \frac{TP}{TP + FN}$$

### 3.3 F-measure and G-measure

F-measure is also known as F-value which uses both precision score and recall score of a classifier. F-measure is another commonly used metric in classification settings. F-measure is calculated using a weighted harmonic mean between precision and recall. For the classification of positive instances, it helps to understand the tradeoff between correctness and coverage [5]. The general formula for calculating F-measure is given below.

$$F_\beta = (1 + \beta^2) * \frac{precision * recall}{(\beta^2 * precision) + recall}$$

In the above formulation, the importance of each term can be provided using different values for $\beta$. Most commonly used value for $\beta$ is 1 which is known as F-1 measure. G-measure is similar to that of F-measure, but it uses geometric mean instead of harmonic mean.

$$F_1 = 2 * \frac{precision * recall}{precision + recall}$$

$$G - measure = \sqrt{precision * recall}$$

In Python, F1-scores can be calculated using "f1_score()" function from "sklearn.metrics" [17], and G-measure scores can be calculated using the "geometric_mean_score()" function from the "imblearn.metrics" library [19].

### 3.4 ROC Curve and AUC

ROC (Receiver Operating Characteristic) curve is used for evaluating the performance of a classifier. It is plotted by calculating False Positive Rate (FPR) on the x-axis against True Positive Rate (TPR) on the y-axis. A range of thresholds from 0 to 1 is defined for a classifier to perform classification. For every point, FPR and TPR are plotted against each other. An example of a ROC curve is shown in Figure 2.

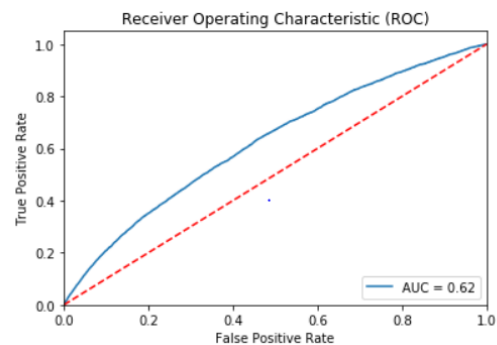

**Figure 2:** ROC Curve and AUC

The ROC's curve top left corner represents good classification, while lower right corner represents poor classification. A classifier is said to be a good classifier if it reaches the top left corner [5]. The diagonal in the plot represents random guessing. If ROC curve of any classifier is below the diagonal, then that classifier is performing poorer than random guessing [5] which entirely defeats the purpose.

Therefore, it is expected that the ROC curve should always be in the upper diagonal. ROC curve helps by providing a graphical representation of a classifier, but it is always a good idea to calculate a numerical score for a classifier. It is a common practice to calculate the Area under the Curve (AUC) for the ROC curve. The AUC value represents a score which is between: 0 to 1. Any classifier whose ROC curve is present in lower diagonal will get an AUC score of less than 0.5. Similarly, the ROC curve which is present in upper diagonal will get AUC scores higher than 0.5. An ideal classifier will get an AUC score of 1 which will touch the upper left corner of the plot. Python provides a function "roc_curve()" to get FPR, TPR, and thresholds for ROC. AUC scores for a classifier can be calculated using the "auc()" function for which users need to provide FPR and TPR of a classifier. Both of these functions can be imported from "*sklearn.metrics*" in Python.

*Note: In some literatures it is mentioned that precision-recall curve and cost measures (cost matrix and cost sensitive curves) can be good performance measures in the case of imbalanced data. Interested readers can refer following resources to get more information about those topics.*

- *N.V. Chawla, "Data mining for imbalanced datasets: An overview", In Data mining and knowledge discovery handbook, pp. 875-886, 2009*

- *H. He, and E.A. Garcia, "Learning from imbalanced data", IEEE Transactions on Knowledge & Data Engineering, vol. 9, pp.1263-1284, 2008*

- *F. Alberto, S. García, M. Galar, R. Prati, B. Krawczyk, and F. Herrera, "Learning from imbalanced data sets", Springer, 2018.*

In this sub-section different statistical assessment metrics applied on "Porto Seguro's Safe Driver Prediction" dataset. This data contains information which can be useful for predicting if a driver will file an insurance claim next year or not. The source for this dataset is www.kaggle.com, more information can be found in reference [16]. In the mentioned dataset, there are 59 columns and around 595k rows. Every row represents details of a policy holder and the target column shows a claim was filed (indicated by '1') or not (indicated by '0'). Before applying different statistical assessment measures, feature selection is performed on all the 59 columns and this resulted into the selection of 22 columns: the name of these 22 columns are - 'target', 'ps_ind_01', 'ps_ind_02_cat', 'ps_ind_03', 'ps_ind_04_cat', 'ps_ind_05_cat', 'ps_ind_06_bin', 'ps_ind_07_bin', 'ps_ind_08_bin', 'ps_ind_09_bin', 'ps_ind_15', 'ps_ind_16_bin', 'ps_ind_17_bin', 'ps_reg_01', 'ps_reg_02', 'ps_reg_03', 'ps_car_03_cat', 'ps_car_07_cat', 'ps_car_11', 'ps_car_12', 'ps_car_13', 'ps_car_15'.

In the "Porto Seguro's Safe Driver Prediction" dataset, there are 573k rows, with target 0, and 21k rows having target 1. It means that there are only 21k people who filled a claim out of 595k drivers. Nonetheless, we can easily see the imbalance in this dataset by observing these numbers. The degree of class imbalance for this dataset using the formula given in section 2 can be calculated as: *26.44*. The code which is used in Python [20] for calculating Imbalanced Ratio is given below.

```python
# Importing libraries
from sklearn.linear_model import LogisticRegression
from sklearn import metrics
from sklearn.model_selection import train_test_split
from sklearn.metrics import accuracy_score

X_train, X_test, y_train, y_test = train_test_split(X, y, test_size=0.3, random_state=0)
logreg = LogisticRegression()
logreg.fit(X_train, y_train)

LogisticRegression(C=1.0, class_weight=None, dual=False, fit_intercept=True,
          intercept_scaling=1, max_iter=100, multi_class='ovr', n_jobs=1,
          penalty='l2', random_state=None, solver='liblinear', tol=0.0001,
          verbose=0, warm_start=False)

y_pred = logreg.predict(X_test)
print('Accuracy of logistic regression classifier on test set: {:.2f}'.format(logreg.score(X_test, y_test)))
Accuracy of logistic regression classifier on test set: 0.96
```

**Figure 3:** Calculation of Imbalanced Ratio

```python
from sklearn.metrics import confusion_matrix
confusion_matrix = confusion_matrix(y_true=y_test, y_pred=y_pred)
print('Confusion matrix:\n', confusion_matrix)

Confusion matrix:
 [[172085      0]
  [  6479      0]]

from sklearn.metrics import precision_score
from sklearn.metrics import recall_score
from  imblearn.metrics import geometric_mean_score
from sklearn.metrics import f1_score

print('Precision score:', precision_score(y_pred, y_test))
print('Recall score:', recall_score(y_pred, y_test))
print('G-mean score:', geometric_mean_score(y_pred, y_test))
print('F1 score:', f1_score(y_pred, y_test))

Precision score: 0.0
Recall score: 0.0
G-mean score: 0.0
F1 score: 0.0
```

**Figure 4:** Implementation of Logistic Regression and Accuracy of the Model

**Figure 5:** Calculation of Confusion Matrix and Other Statistical Assessment Metrics

Further, Logistic Regression is used to perform classification on the dataset, and it can be observed that the accuracy of the model is **96%**. In the next step, confusion matrix and other statistical assessment metrics were calculated in Python.

```python
from sklearn.metrics import roc_auc_score
from sklearn.metrics import roc_curve
from sklearn.metrics import auc
```

```python
logit_roc_auc = roc_auc_score(y_test, y_pred)
fpr, tpr, thresholds = roc_curve(y_test, logreg.predict_proba(X_test)[:,1])
plt.figure()
plt.plot(fpr, tpr, label='AUC = %0.2f' % auc(fpr, tpr))
plt.plot([0, 1], [0, 1],'r--')
plt.xlim([0.0, 1.0])
plt.ylim([0.0, 1.05])
plt.xlabel('False Positive Rate')
plt.ylabel('True Positive Rate')
plt.title('Receiver Operating Characteristic (ROC)')
plt.legend(loc="lower right")
plt.show()
```

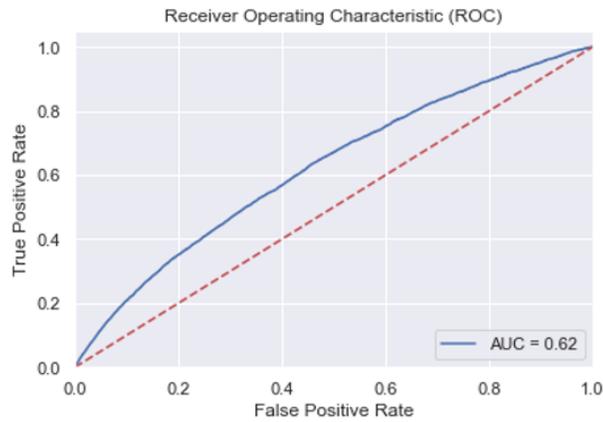

**Figure 6:** The ROC and AUC for the Logistic Regression Model

The results obtained from the analysis can be summarized as follows:

|  |  | Predicted | |
|---|---|---|---|
|  |  | **Negative** | **Positive** |
| **Actual** | **Negative** | 172085 | 0 |
|  | **Positive** | 6479 | 0 |

**Table 2:** Confusion Matrix for the Porto Seguro's Safe Driver Prediction Dataset

| Statistical assessment metrics | Result |
|---|---|
| Accuracy | 0.96 |

| Precision | 0 |
|---|---|
| Recall | 0 |
| F1 score | 0 |
| G-mean score | 0 |

**Table 3:** Other Results for the Porto Seguro's Safe Driver Prediction Dataset

It can be observed from the above results that the model has high quality results, but the model is unable to classify any instance as the positive class. It can also be observed that precision, recall, F1 score, and G-mean score values are 0. Thus, it is a strong indication that balancing the class is critical before making further predictions.

## 4. How to deal with imbalanced data

The previous section explained the risk of using imbalanced data for prediction without preprocessing, which also reflects the critical need to perform preprocessing on the data. This part of the chapter concentrates on data-level methods which are sampling methods. The sampling methods can be divided into three categories: undersampling, oversampling, and hybrid methods. For undersampling and oversampling three different approaches will be explored. Further, these methods will be implemented on Porto Seguro's Safe Driver Prediction dataset, and that will be compared with results from the previous section.

In general, sampling methods modify imbalanced dataset to make it more balanced or make more adequate data distribution for learning tasks [5]. In Python, these sampling methods can be implemented using "imbalanced-learn" library [19]. The link for documentation and installation details of "imbalanced-learn" library can be found in reference [21].

### 4.1 Undersampling

Undersampling methods eliminate the majority class instances in the data to make the data more balanced. In some literature, under-sampling is also called "down-sizing". There are different methods which can be used for undersampling, but three commonly used methods are covered in this sub-section: Random Undersampling, Tomek Links, and Edited Nearest Neighbors (ENN).

#### 4.1.1 Random Undersampling

Random Undersampling is the non-heuristic method for undersampling [5]. It is one of the simplest methods and generally used as a baseline method. In random undersampling, the instances from the negative class or majority class are selected at random, and they are removed until it matches the count of positive class or minority class. This technique selects instances from the majority class randomly and thus, it is

```python
from imblearn.under_sampling import RandomUnderSampler
rus = RandomUnderSampler(random_state=0)
X_resampled, y_resampled = rus.fit_resample(X, y)
len(X_resampled)
```
43388

```python
print(sorted(Counter(y_resampled).items()))
```
[(0, 21694), (1, 21694)]

known as "Random undersampling". The result of this technique will be a balanced data set consisting of an equal number of positive class and negative class examples.

**Figure 7:** Random Undersampling Implementation in Python

Random undersampling can be performed by importing "RandomUnderSampler" from "imblearn.under_sampling". In the above example, an object "rus" is created for "RandomUnderSampler()" and then "fit.resample()" method is used by providing data (as 'x') and labels (as 'y') independently. To reproduce the results "random_state = 0" is used in the "RandomUnderSampler()" function. The result will get stored in "X_resampled" (details of all the columns except the target) and "Y_resampled" (the target column). After performing Random under-sampling, it can be observed that there is an equal number of positive (21,694) and negative class (21,694) instances in the dataset. The results obtained after performing logistic regression on the processed data are summarized as follows.

|  |  | Predicted | |
|---|---|---|---|
|  |  | **Negative** | **Positive** |
| **Actual** | **Negative** | 4115 | 2402 |
|  | **Positive** | 2878 | 3622 |

**Table 4:** Confusion Matrix for Porto Seguro's Safe Driver Prediction Dataset

| **Statistical assessment metrics** | **Result** |
|---|---|
| Accuracy | 0.59 |
| Precision | 0.56 |
| Recall | 0.60 |
| F1 score | 0.58 |
| G-mean score | 0.59 |

**Table 5:** Other Results for Porto Seguro's Safe Driver Prediction Dataset

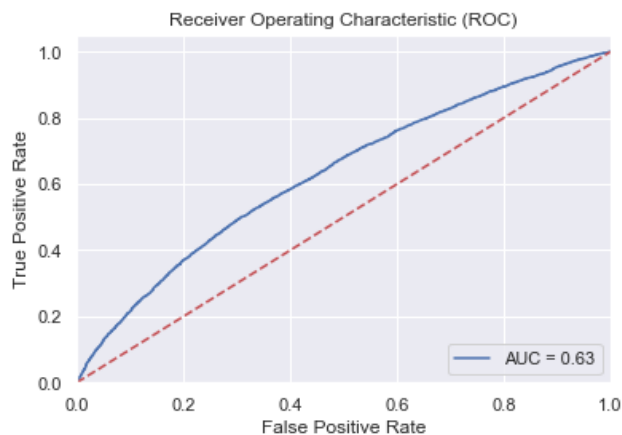

**Figure 8:** ROC and AUC After Performing Random Undersampling

It can be observed after performing Random undersampling the accuracy is 59%, and the model can classify positive class instances. From the results, it can also be seen that precision, recall, F1 score, and G-mean score values are not 0, but it reflects a serious chance to improve the performance of the model.

### 4.1.2 Tomek Link

Tomek Link is a heuristic undersampling technique based on a **distance measure**. In Tomek Link – a link is established based on a distance between instances from two different classes which further used for removing majority class instances [22]. The conceptual working of the Tomek Link is given below and is motivated from Fernández et al. [5].

1) Consider two examples: $A_i$ and $A_j$ where $A_i$ can be represented as $(x_i, y_i)$ and $A_j$ can be represented as $(x_j, y_j)$.
2) The distance between $A_i$ and $A_j$ is D which can be represented as $d(A_i, A_j)$.
3) A pair $(A_i, A_j)$ can said to have Tomek Link if there is not an example $A_l$ such that $d(A_i, A_l) < d(A_i, A_j)$ or $d(A_j, A_l) < d(A_i, A_j)$.
4) After identifying Tomek Links, the instance which belongs from the majority class is removed, and instances from the minority class is kept in the dataset.

```
from imblearn.under_sampling import TomekLinks
tl=TomekLinks(sampling_strategy='majority')
X_resampled_tl, y_resampled_tl = tl.fit_resample(X, y)
len(X_resampled_tl)
```
585471

```
print(sorted(Counter(y_resampled_tl).items()))
```
[(0, 563777), (1, 21694)]

**Figure 9:** Tomek Links Implementation in Python

Tomek Links can be implemented in Python using "TomekLinks()" function from "imblearn.under_sampling". One of the parameters in the function is "sampling_strategy" which can be used for defining sampling strategy. The default sampling strategy is "auto" which means resample all classes but not minority classes. The additional details about the function parameters can be found in the documentation. After performing Tomek Link, it can be observed that there are still less positive (21,694) class instances as compared to the negative class (563,777) instances. The results obtained after performing logistic regression on the processed data are summarized as follows.

|  |  | Predicted | |
|---|---|---|---|
|  |  | **Negative** | **Positive** |
| **Actual** | **Negative** | 169215 | 0 |
|  | **Positive** | 6427 | 0 |

**Table 6:** Confusion Matrix for Porto Seguro's Safe Driver Prediction Dataset

| **Statistical assessment metrics** | **Result** |
|---|---|
| Accuracy | 0.96 |
| Precision | 0 |
| Recall | 0 |
| F1 score | 0 |
| G-mean score | 0 |

**Table 7:** Other Results for Porto Seguro's Safe Driver Prediction Dataset

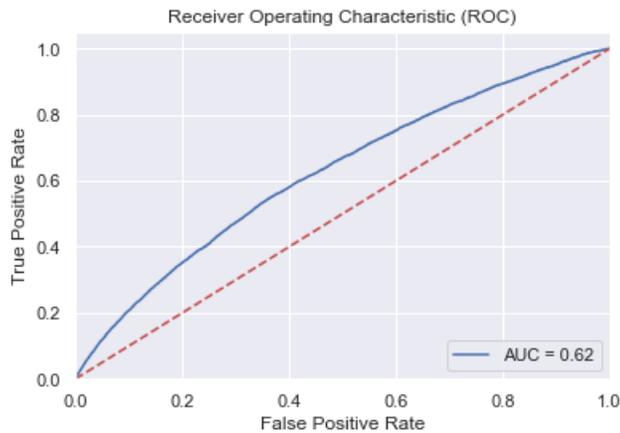

**Figure 10:** ROC and AUC After Performing Tomek Links

After performing Tomek Link, the observed accuracy is **96%**, but the model is unable to classify positive class instances. It can also be noted that precision, recall, F1 score, and G-mean score values are 0. Thus, in this scenario, Tomek links are not a suitable strategy for balancing the class distribution.

### 4.1.3   Edited Nearest Neighbors (ENN)

ENN is the third method which is a part of undersampling category. In ENN, the instances which belong from the majority class are removed based on their K-nearest neighbors [23]. For example, if 3 neighbors are considered, then the majority class instance will get compared with their closest 3 neighbors. If most of their neighbors are minority classes, then that instance will get removed from the dataset.

```python
from imblearn.under_sampling import EditedNearestNeighbours
enn=EditedNearestNeighbours(sampling_strategy='majority')
X_resampled_enn, y_resampled_enn = enn.fit_resample(X, y)
len(X_resampled_enn)
```
538120

```python
print(sorted(Counter(y_resampled_enn).items()))
```
[(0, 516426), (1, 21694)]

**Figure 10:** ENN Implementation in Python

ENN can implemented in Python using "EditedNearestNeighbours()" function from "imblearn.under_sampling". It also has a parameter "sampling_strategy" which can be used for defining sampling strategy. The default sampling strategy is "auto", which means resampling all classes but not minority classes. In addition to that, users can also set "n_neighbors" parameter, which can be used to set the size of the neighborhood to compute nearest neighbors. The default value for "n_neighbors" parameter is 3. Results obtained after performing logistic regression on the processed data are summarized as follows.

**Predicted**

|        |          | Negative | Positive |
|--------|----------|----------|----------|
| Actual | Negative | 155003   | 0        |
|        | Positive | 6433     | 0        |

**Table 6:** Confusion Matrix for Porto Seguro's Safe Driver Prediction Dataset

| Statistical assessment metrics | Result |
|---|---|
| Accuracy | 0.96 |
| Precision | 0 |
| Recall | 0 |
| F1 score | 0 |
| G-mean score | 0 |

**Table 7:** Other Results for Porto Seguro's Safe Driver Prediction Dataset

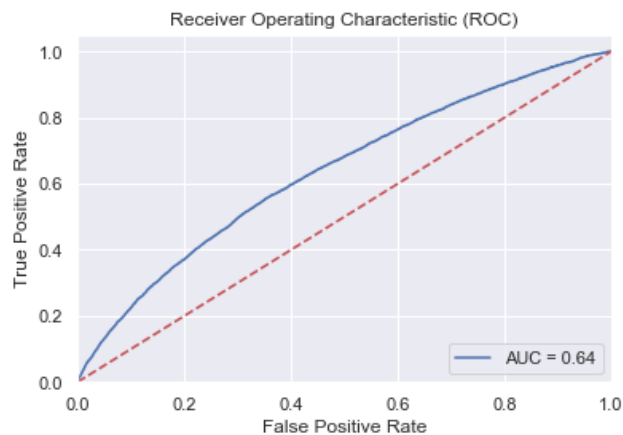

**Figure 8:** ROC and AUC After Performing ENN

The results indicate that the accuracy is **96%**, but the model is unable to classify positive class instances, same as Tomek link. Therefore, in this scenario, ENN is also not a suitable strategy for balancing the class distribution.

> *Note: There are some other methods for under-sampling such as Class Purity Maximization (CPM), Undersampling based on Clustering (SBC), NearMiss approaches as well as some advanced techniques using evolutionary algorithms. Interested readers can get more information about these techniques in F. Alberto, S. García, M. Galar, R. Prati, B. Krawczyk, and F. Herrera, "Learning from imbalanced data sets", Springer, 2018.*

### 4.2 Oversampling

This sub-section focuses on oversampling methods that work oppositely as compared to undersampling. In oversampling, the goal is to increase the count of minority class instances to match it with the count of majority class instances. Thus, oversampling is "upsizing" the minority class. In this section, three commonly used oversampling methods are covered: Random Oversampling, Synthetic Minority Oversampling Technique (SMOTE), and Adaptive Synthetic Sampling (ADASYN).

### 4.2.1 Random Oversampling

Random oversampling is a non-heuristic method under the category of oversampling methods [5]. In Random oversampling, instances from the minority class are selected at random for replication which results in a balanced class distribution. In this method, the minority class instances are chosen at random, and therefore, this method is called "Random".

```
from imblearn.over_sampling import RandomOverSampler
ros = RandomOverSampler(random_state=0)
X_resampled_ros, y_resampled_ros = ros.fit_resample(X, y)
len(X_resampled_ros)
```
1147036

```
print(sorted(Counter(y_resampled_ros).items()))
```
[(0, 573518), (1, 573518)]

**Figure 11:** Random Oversampling Implementation in Python

Random Oversampling can be performed in Python using "RandomOverSampler()" from "imblearn.over_sampling". The function supports different parameters for resampling, but the default parameter is "auto" which is equivalent to not to resample majority class. More information about other parameters of "RandomOverSampler()" can be found in the documentation. After performing Random Oversampling, it can be observed that there is an equal number of positive (573,518) and negative class (573,518) instances in the dataset. The results obtained after performing logistic regression on the processed data are summarized as follows.

|  |  | Predicted | |
|---|---|---|---|
|  |  | **Negative** | **Positive** |
| **Actual** | **Negative** | 108341 | 63542 |
|  | **Positive** | 77842 | 94386 |

**Table 8:** Confusion Matrix for Porto Seguro's Safe Driver Prediction Dataset

| Statistical assessment metrics | Result |
|---|---|
| Accuracy | 0.59 |
| Precision | 0.54 |
| Recall | 0.60 |
| F1 score | 0.59 |
| G-mean score | 0.57 |

**Table 9:** Other Results for Porto Seguro's Safe Driver Prediction Dataset

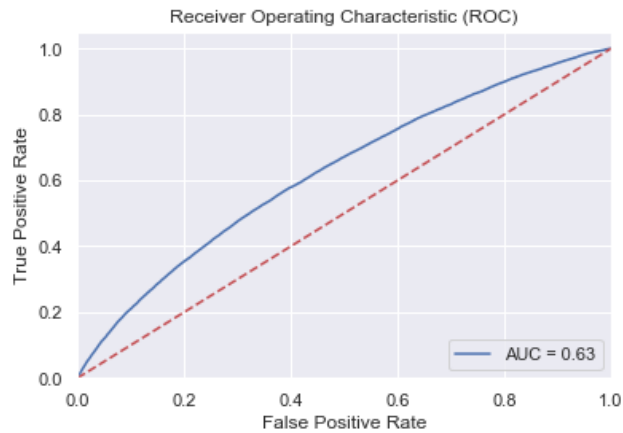

**Figure 12:** ROC and AUC After Performing Random Oversampling

It can be observed that the accuracy is **59%** and the model can classify positive class instances after performing Random oversampling. It can also be noted that precision, recall, F1 score, and G-mean score values are less than or equal to **0.6**. These results indicate that Random Oversampling provides more meaningful results as compared to Tomek Link and ENN.

### 4.2.2 Synthetic Minority Oversampling Technique (SMOTE)

SMOTE is another popular method for performing oversampling. The previous method - Random Oversampling can led to overfitting because of randomly replicating the instances of the minority class [24]. Therefore, to overcome this problem of overfitting, Chawla et al. [24] developed a new technique for generating minority class instances synthetically. In SMOTE, new instances are created based on interpolation between several minority class instances that lie together [5]. This allows SMOTE to operate in feature space rather than to operate in data space [24]. The detailed working of the SMOTE method along with the discussion can be found in "SMOTE: Synthetic Minority Oversampling Technique" [24].

```python
from imblearn.over_sampling import SMOTE
sm = SMOTE(random_state=42,k_neighbors=5)
X_resampled_sm, y_resampled_sm = sm.fit_resample(X, y)
len(X_resampled_sm)
```
1147036

```python
print(sorted(Counter(y_resampled_sm).items()))
```
[(0, 573518), (1, 573518)]

**Figure 13:** SMOTE Implementation in Python

In Python, SMOTE can be implemented to perform oversampling using the "SMOTE()" function from "imblearn.over_sampling". SMOTE method is based on the number of neighbors, and it can be defined in the "SMOTE()" function using "k_neighbors" parameter. The default number of neighbors are five, but users can tweak this parameter. Users can also define a strategy for resampling like other functions and default strategy is "auto" which means not to target majority class. After performing SMOTE, it can be observed that there is an equal number of positive (573,518) and negative class (573,518) instances in the dataset. The results obtained after performing logistic regression on the processed data are summarized as follows.

|        |          | Predicted |          |
|--------|----------|-----------|----------|
|        |          | Negative  | Positive |
| Actual | Negative | 103484    | 68772    |
|        | Positive | 74515     | 97340    |

**Table 10:** Confusion Matrix for Porto Seguro's Safe Driver Prediction Dataset

| Statistical assessment metrics | Result |
|---|---|
| Accuracy | 0.58 |
| Precision | 0.57 |
| Recall | 0.59 |
| F1 score | 0.58 |
| G-mean score | 0.58 |

**Table 11:** Other Results for Porto Seguro's Safe Driver Prediction Dataset

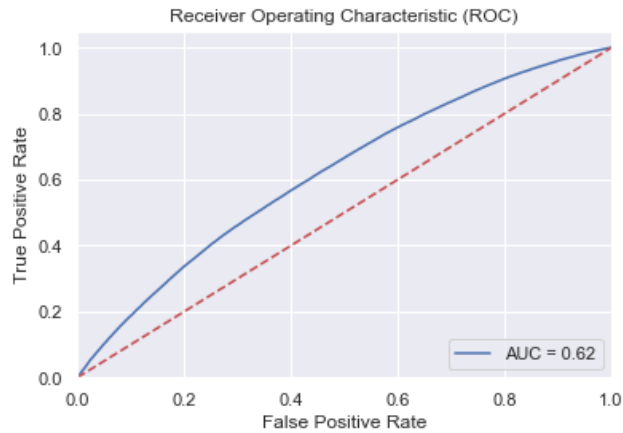

**Figure 14:** ROC and AUC After Performing Random Oversampling

It can be observed from the above results that the accuracy is **58%**. Also, precision, recall, F1 score, and G-mean score values are less than **0.6**.

### 4.2.3 Adaptive Synthetic Sampling (ADASYN)

ADASYN is one of the popular extensions of the SMOTE which is developed by Haibo et al. [26]. There are more than 90 methods which are based on SMOTE and for the complete list readers can refer to [5]. In ADASYN the minority examples are generated based on their density distribution – more synthetic data is generated from minority class samples that are harder to learn as compared to those minority samples that are easier to learn [26]. The two objectives behind ADASYN technique are– reducing the bias and deploying adaptively learning. The detailed workings of the ADASYN method along with the discussion can be found in "ADASYN: Adaptive Synthetic Sampling Approach for Imbalanced Learning" by Haibo et al. [26].

```
from imblearn.over_sampling import ADASYN
ad = ADASYN(random_state=42,n_neighbors=5)
X_resampled_ad, y_resampled_ad = ad.fit_resample(X, y)
len(X_resampled_ad)
```
1152015

```
print(sorted(Counter(y_resampled_ad).items()))
```
[(0, 573518), (1, 578497)]

**Figure 15:** ADASYN implementation in Python

In Python, ADASYN can be implemented using the "ADASYN()" function from "imblearn.over_sampling". In "ADASYN()" there is a parameter "n_neighbors" for providing a number of nearest neighbors for creating synthetic samples and the default value for this parameter is 5. In ADASYN, randomization of the algorithm can be controlled using the "random_state" parameter. After performing ADASYN, it can be observed that the count of positive (573,518) and negative class (578,497) instances are very close. The results obtained after performing logistic regression on the processed data are summarized as follows.

|  |  | Predicted | |
|---|---|---|---|
|  |  | **Negative** | **Positive** |
| **Actual** | **Negative** | 101158 | 70906 |
|  | **Positive** | 73534 | 100007 |

**Table 12:** Confusion Matrix for Porto Seguro's Safe Driver Prediction Dataset

| Statistical assessment metrics | Result |
|---|---|
| Accuracy | 0.58 |
| Precision | 0.58 |
| Recall | 0.59 |
| F1 score | 0.58 |
| G-mean score | 0.58 |

**Table 13:** Other Results for Porto Seguro's Safe Driver Prediction Dataset

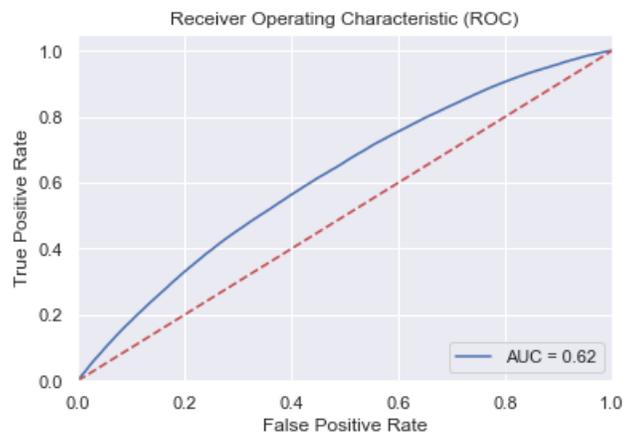

**Figure 16:** ROC and AUC After Performing ADASYN

It can be observed that the accuracy is **58%**. It can also be noted that precision, recall, F1 score, and G-mean score values are less than **0.6**. The AUC value obtained after performing analysis is **0.62**.

### 4.3 Hybrid Methods

The most popular undersampling and oversampling methods are reviewed and implemented in previous sub-sections. In undersampling, there is a removal of majority class instances, while in oversampling, there is a creation of minority class instances for making a class distribution equal. Based on the same theme, hybrid methods are a combination of undersampling and oversampling methods. Thus, hybrid methods help to achieve the balance between removing majority class instances and creating minority class instances. In hybrid methods, SMOTE is one of the popular techniques used for oversampling which can be used with Tomek Link or ENN for undersampling.

Python also provides the functionality to apply two hybrid methods – "SMOTEENN()" (SMOTE + ENN) and "SMOTETomek()" (SMOTE + Tomek Link) from "imblearn.combine" in Python. More details about parameters and implementation can be found in reference [21].

### 5. Other Methods

In the previous section, data-level methods such as – undersampling, oversampling, and hybrid methods are discussed. The data-level methods primarily perform sampling to deal with the problem of imbalance in our dataset. In addition to data-level methods, there are also algorithmic-level approaches and ensemble-based approaches. Algorithmic-level approaches are more focused on modifying the classifier instead of modifying the dataset [5]. These methods require to have a good understanding of the classifier methods and how they are affected by an imbalanced dataset. Some of the commonly used techniques in algorithmic-level approaches are: kernel-based approaches, weighted approaches, active learning, and one-class learning. Ensemble-based approaches combine several classifiers at the final step by combining their outputs. Most commonly used ensemble-based approaches are: bagging, and boosting. The scope of algorithmic-level approaches and ensemble-based approaches are not part of this chapter, and interested readers can get a useful review of these methods in "Learning from imbalanced data sets" by Fernández et al. [5].

**Conclusion**

This chapter introduced the problem of data imbalance and different data-level methods to deal with this problem. The chapter began with the definition of imbalance, positive class, negative class and then focused on the main causes of imbalance which are: class overlap and small disjoints. Afterwards, different statistical assessment metrics were discussed, implemented, and summarized after performing logistic regression on the "Porto Seguro's Safe Driver Prediction" dataset. The most important part of this chapter was the discussion and comparison of different data-level methods. For the data-level methods, it can be observed that for this scenario oversampling methods performed better as compared to undersampling. In the case of undersampling, only Random under-sampling can classify positive class instances after processing the data. While all the oversampling methods which are discussed in this chapter are able to produce better results. Results from oversampling methods indicate that all the methods were able to provide a similar type of results (accuracy, precision, recall, F1 score, G-mean score, and AUC score).

It is critical to note that there is no single method which can be suitable for all problems. In this scenario, oversampling methods are outperforming undersampling methods, but the critical step is to use multiple methods for pre-processing the data and then compare the results for selecting the best possible method based on the type of the data as well as by considering the scenario of the problem. The analysis performed in this chapter also points out the importance of feature engineering. Use of a combination of different features and/or selection of features before applying these methods can generate better results. In the real world, it is also essential to understand the meaning of the results and how these results will help to solve a business problem. The use of undersampling methods remove the majority class samples from the dataset which can be problematic in many cases as every record in the dataset may denote something significant. Similarly, the use of oversampling methods generates samples from the data which can result in overfitting of the model or creating irrelevant records in the dataset. Therefore, caution should always be taken while using data-level approaches especially while dealing with sensitive information. As this book aims to present, technical and statistical methods presented in this chapter are essential in avoiding blind spots and ensuring that the correct representation of facts, data, and the truth are being extracted from the data; that notion leads to better data science, and a data democracy.